\documentclass[a4paper]{styles/svproc}

\usepackage{url}

\usepackage{picinpar}
\usepackage{floatflt}

\usepackage{stfloats}
\usepackage{makecell}
\usepackage{color}
\usepackage{algorithm}
\usepackage{algorithmic}
\usepackage{graphicx}
\usepackage{amssymb}
\usepackage{amsfonts}
\usepackage{amsmath}
\usepackage{hyperref}
\usepackage{multirow}
\hypersetup{
    colorlinks=true,
    linkcolor=blue,
    urlcolor=blue,
    breaklinks=true,
    citecolor=green,
    }

\usepackage{balance}
\usepackage{cite}
\usepackage[caption=false,font=footnotesize,listofformat=subsimple,labelformat=simple]{subfig}

\makeatletter
\renewcommand\paragraph{\@startsection{paragraph}{4}{\z@}%
    {3.25ex \@plus1ex \@minus.2ex}%
    {-1em}%
    {\normalfont\normalsize\bfseries}}
\makeatother

\begin{document}
\mainmatter              
\title{Learning from Few Demonstrations with Frame--Weighted Motion Generation}
\titlerunning{Learning from Few Demonstrations}  

\author{Jianyong Sun\inst{1} \and
Jens Kober\inst{1} \and Michael Gienger\inst{2} \and Jihong Zhu\inst{1, 3}}
\authorrunning{J. Sun et al.} 
%
%
\institute{Delft University of Technology, 2628CD Delft, The Netherlands\\
\email{sunjy711@gmail.com}, \email{\{J.Kober,J.Zhu-3\}@tudelft.nl}
\and
Honda Research Institute Europe, 63073 Offenbach/Main, Germany\\
\email{michael.gienger@honda-ri.de}
\and
University of York, York YO10 5DD, United Kingdom\\
\email{jihong.zhu@york.ac.uk}
}

\maketitle              
\setcounter{footnote}{0}

\begin{abstract}
Learning from Demonstration (LfD) enables robots to acquire versatile skills by learning motion policies from human demonstrations. It endows users with an intuitive interface to transfer new skills to robots without the need for time-consuming robot programming and inefficient solution exploration. During task executions, the robot motion is usually influenced by constraints imposed by environments. In light of this, task-parameterized LfD (TP-LfD) encodes relevant contextual information into reference frames, enabling better skill generalization to new situations. However, most TP-LfD algorithms typically require multiple demonstrations across various environmental conditions to ensure sufficient statistics for a meaningful model. It is not a trivial task for robot users to create different situations and perform demonstrations under all of them. Therefore, this paper presents a novel algorithm to learn skills from few demonstrations. By leveraging the reference frame weights that capture the frame importance or relevance during task executions, our method demonstrates excellent skill acquisition performance, which is validated in real robotic environments\footnote{The video of the experiments is available at \href{https://youtu.be/JpGjk4eKC3o}{https://youtu.be/JpGjk4eKC3o}.}.
\end{abstract}

\section{Introduction} \label{sec: introduction}
Learning from Demonstration (LfD) aims to encode versatile skills from human demonstrations \cite{ravichandar_recent_2020}. Popular movement encoding methods include Dynamic Movement Primitive (DMP) \cite{ijspeert_dynamical_2013}, Probabilistic Movement Primitive (ProMP) \cite{paraschos_probabilistic_2013}, Gaussian Mixture Model (GMM) \cite{calinon_learning_2007}, etc. To endow robots with the ability to adapt their learned skills to various situations, many LfD algorithms integrate these methods with task parameters that can describe task situations \cite{forte_-line_2012, pervez_learning_2018, osa_guiding_2017}. However, most of them formulate the motion retrieval from task parameters as a regression problem, hence exhibiting limited extrapolation ability. To improve that, \cite{calinon_tutorial_2016} proposes the Task-Parameterized LfD (TP-LfD) method by parameterizing the description of task situations as reference frames. Taking the movement task in Fig. \ref{fig: intro frame} as an example, since the poses of the two boxes determine the task situation, it can be fully described by two reference frames attached to each respective box. As a typical example of TP-LfD methods, Task-Parameterized Gaussian Mixture Model (TP-GMM) employs GMM to encode demonstrations within different local reference frames and fuses local trajectory distributions in a new situation, yielding reliable generalization performance \cite{calinon_tutorial_2016}.

\begin{floatingfigure}[r]{0.25\textwidth}
    \centering
    \includegraphics[width=0.25\textwidth]{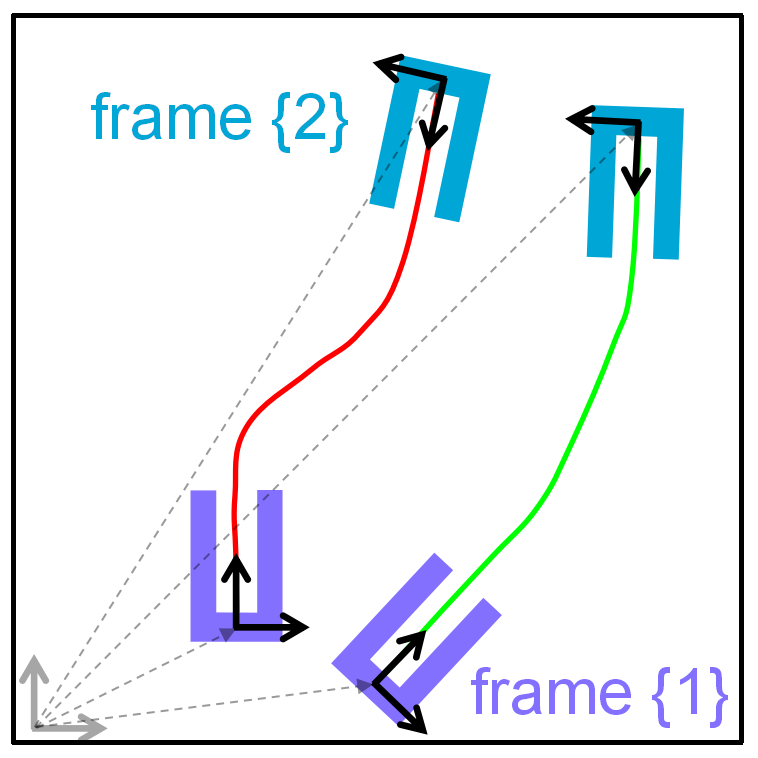}
    \caption{Two situations of a simulated movement task from the purple U-shape box to the blue one.}
    \label{fig: intro frame}
\end{floatingfigure}

To obtain enough training data, existing TP learning methods typically demand an extensive collection of diverse demonstrations in various environmental situations. However, this requirement can be challenging to fulfill in practice due to resource constraints, especially for long-horizon tasks. If the training demonstrations are few, these methods often struggle to extract adequate information from the data, hence showing poor generalization performance. Taking advantage of the TP formulation, we here propose a novel learning from few demonstrations method by leveraging task frame relevance instead of directly augmenting the original demonstrations, while the latter is conventionally done \cite{zhu_learning_2022}.

Varying throughout a task trajectory, the frame relevance represents the individual influence or importance of a reference frame concerning a portion of the task \cite{huang_generalized_2018}. For example, in the scenario shown in Fig. \ref{fig: intro frame}, as the trajectory approaches the blue box, it is more influenced by frame $\{2\}$, while frame $\{1\}$ becomes less relevant. \cite{huang_generalized_2018} manually specifies relevance weights for reference frames based on human prior knowledge. The methods presented in \cite{alizadeh_learning_2014} and \cite{sena_improving_2019} rely heavily on data to calculate the weights, rendering them not applicable to our case where only few demonstrations are available for training.

The main contributions of this paper are as follows:
\begin{enumerate}   
    \item We formalize a motion generation method that utilizes the relevance weights of reference frames to transform demonstrated trajectories.
    \item We formulate the computation of frame relevance weights as an optimization problem, which can be solved with only a limited number of demonstrations.
    \item We show that our frame--weighted motion generation method can not only allow for direct application to generalization but also augment the original training dataset to achieve policy improvements for traditional TP-LfD methods.
\end{enumerate}

\section{Technical Approach} \label{sec: approach}
\subsection{Problem Statement} \label{sec: problem statement}
We consider two local reference frames and employ the rotation matrix $\pmb A_{j}$ and translation vector $\pmb b_{j}$ of each frame $\{ j \}$ with respect to the global coordinate system as task parameters ($j = 1, 2$)\footnote{Our method is not limited to cases with two reference frames. See Sect. \ref{sec: multi-frame task solution}.}. The training dataset $\{\{\pmb \xi_{t, m}\}_{t=0}^{T_m}\}_{m=1}^M$ consists of $M$ demonstrations, each with the time length $T_m$. Their associated reference frames are denoted by $\{\{ \pmb A_{j, m}, \pmb b_{j, m}\}_{j=1}^2 \}_{m=1}^M$. For the $m$-th demonstration, $\pmb \xi_{t, m}$ represents a trajectory point recorded at the time step $t$. 

Since the frame relevance varies during task executions, we begin by defining a progress scalar related to the task trajectory to describe the motion process. In our case, we use the relative distance index $d_{t, m}$ of each data point $\pmb \xi_{t, m}$
\begin{equation} \label{eq: dist}
    \begin{split}
    d_{t, m} = \left \{
        \begin{array}{ll}
        0, & \, t=0 \\
        \frac{\sum_{i=0}^{t-1} \lVert \pmb \xi_{i, m} - \pmb \xi_{i+1, m} \rVert }{\sum_{i=0}^{T_m-1} \lVert \pmb \xi_{i, m} - \pmb \xi_{i+1, m} \rVert }, & \, t>0 
        \end{array}
        \right.
    \end{split}
\end{equation}
where $\lVert \cdot \rVert$ represents the Euclidean distance between two consecutive trajectory points. Along a trajectory, the index $d_{t, m}$ starts from 0 and reaches 1 at the end.

We then define the relevance weights $f_j(d) \in (0, 1)$ of frame $\{j\}$ as a weighted summation of $Q$ Radial Basis Functions (RBFs)
\begin{equation} \label{eq: initial weight definition}
    f_j(d) = \pmb{\Phi}^T \pmb{\omega}_j,
\end{equation}
where the vector $\pmb{\Phi}$ is defined as $[\phi_1(d), \dots, \phi_Q(d)]^T$ whose entries each denote a RBF regarding the relative distance index $d$ and $\pmb{\omega}_j \in \mathbb{R}^Q$ denotes the basis function weight vector. By doing so, we transform the problem of finding frame relevance weights as the computation of basis function weights. 

Inspired by the method to calculate frame relevance weights in \cite{sena_improving_2019}, we formulate the determination of basis function weights $\pmb{\omega}_j$ for frame $\{j\}$ as an optimization problem. In Sect. \ref{sec: reference trajectory transformation}, we first introduce how to utilize frame relevance weights for reference trajectory transformation, underlying the construction of the objective function. We then present the solution of the optimization problem in Sect. \ref{sec: frame relevance weights optimization} and the formation of our frame--weighted motion generation method in Sect. \ref{sec: skill generalization}. Finally, we illustrate the application of our approach to the task that requires multiple reference frames to describe the situation in Sect. \ref{sec: multi-frame task solution}.

\subsection{Reference Trajectory Transformation} \label{sec: reference trajectory transformation}
\begin{figure}[!t]
    \centering
    \subfloat[Demonstr.]{\includegraphics[width=0.19\textwidth]{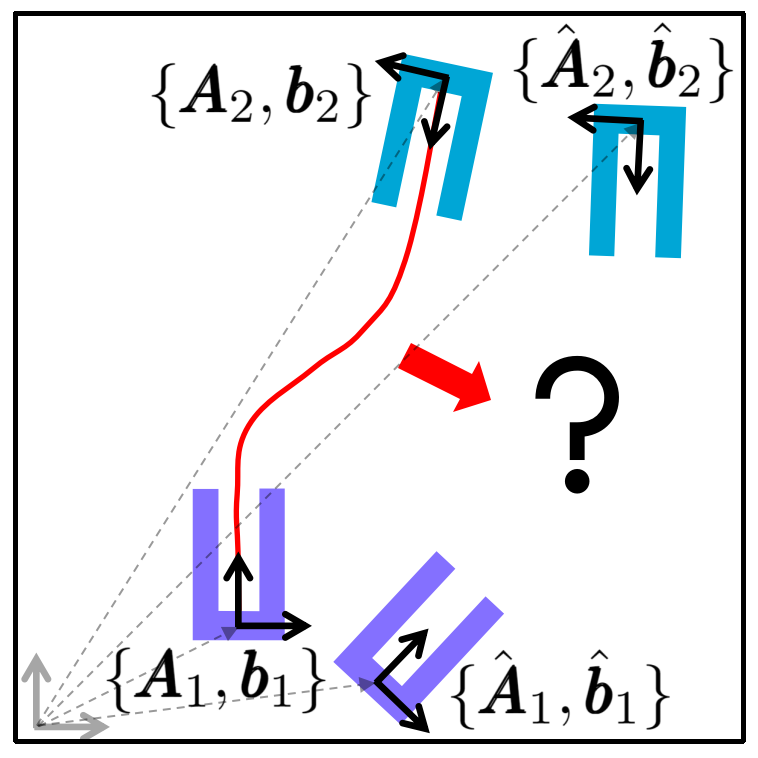}
    \label{fig: method step 0}}
    \hfil
    \subfloat[Step 1]{\includegraphics[width=0.19\textwidth]{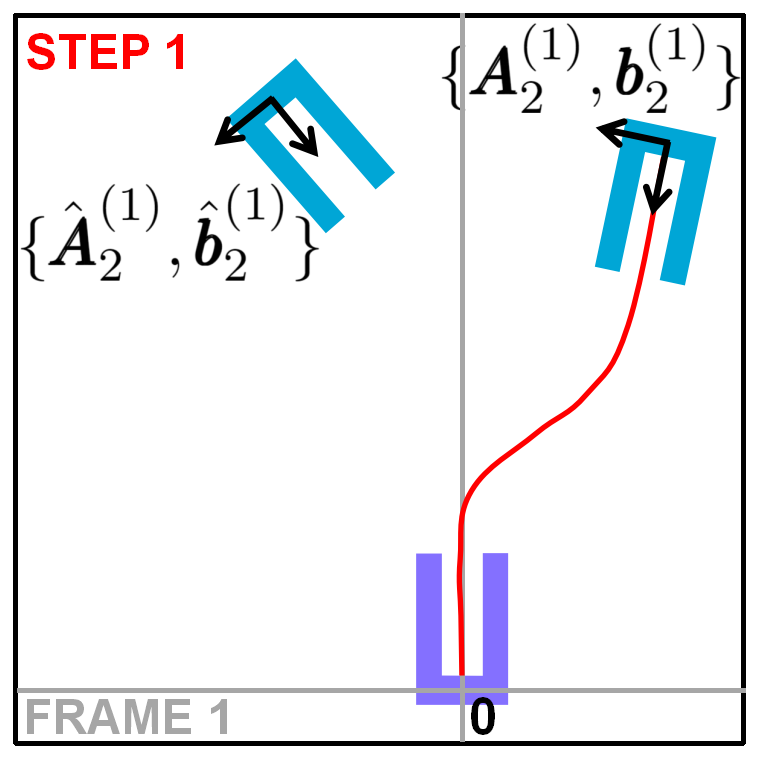}
    \label{fig: method step 1}}
    \hfil
    \subfloat[Step 2]{\includegraphics[width=0.19\textwidth]{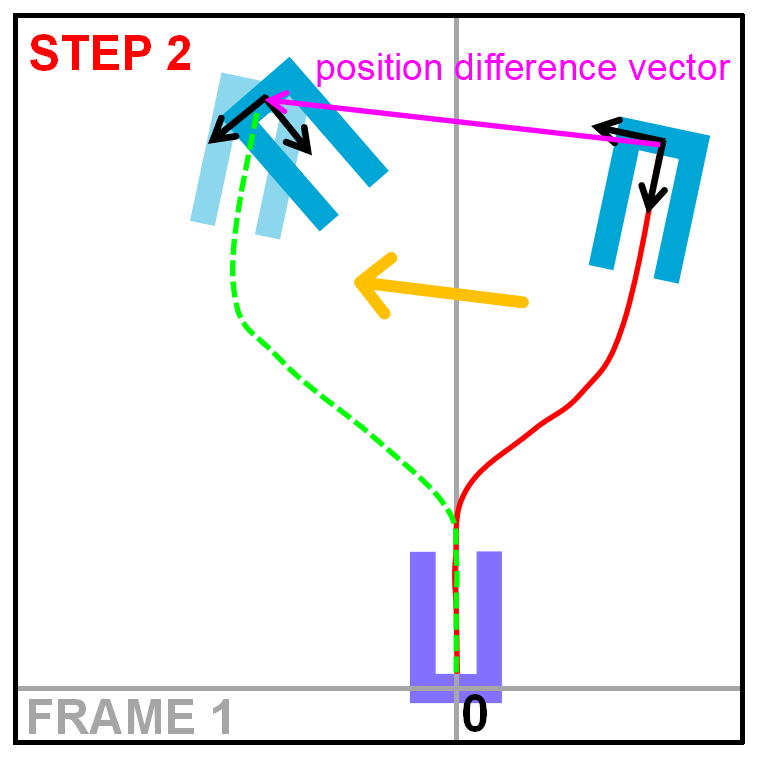}
    \label{fig: method step 2}}
    \hfil
    \subfloat[Step 3]{\includegraphics[width=0.19\textwidth]{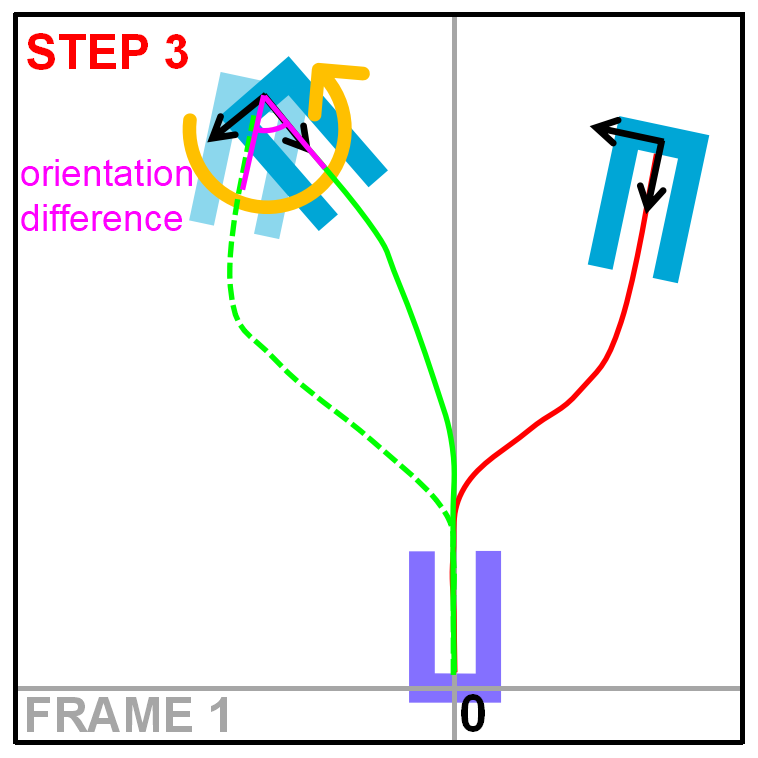}
    \label{fig: method step 3}}
    \hfil
    \subfloat[Step 4]{\includegraphics[width=0.19\textwidth]{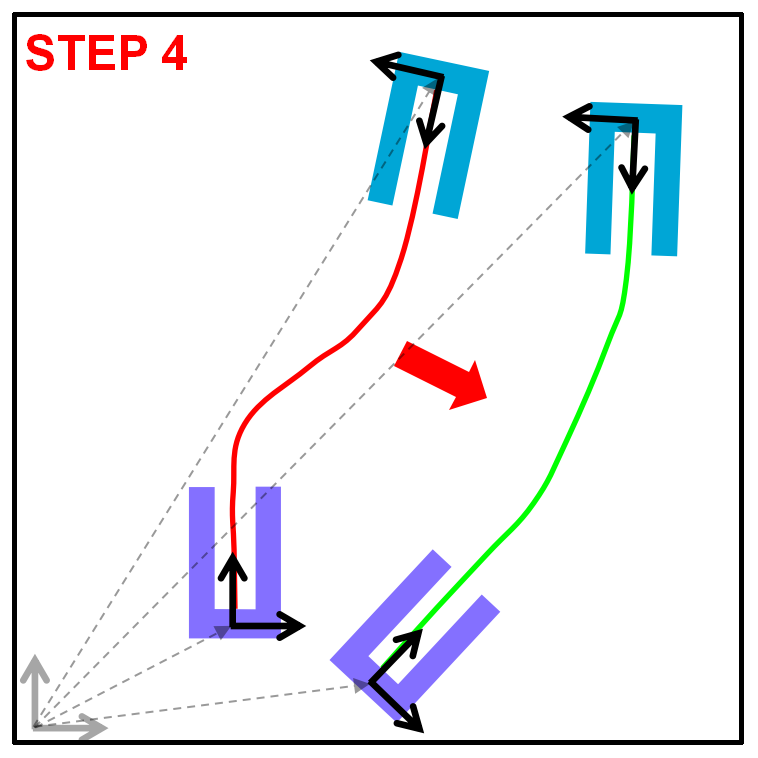}
    \label{fig: method step 4}}
    
    \caption{Reference trajectory transformation using frame relevance weights.}
    \label{fig: method steps}
\end{figure}

This section details how to take a demonstration as the reference trajectory and transform it for motion generation in a different situation. As shown in Fig. \ref{fig: method step 0}, we express the demonstration as $\{\pmb \xi_{t}\}_{t=0}^{T}$ and its associated reference frames as $\{ \pmb A_{j}, \pmb b_{j}\}_{j=1}^2$. We define the new situation using reference frames $\{ \hat{\pmb A}_{j}, \hat{\pmb b}_{j}\}_{j=1}^2$. The four steps to be described correspond to Fig. \ref{fig: method step 1}--\ref{fig: method step 4}.

\vspace{-4mm}
\subsubsection{Step 1: Projection to Local.} We first project each frame $\{2\}$ from the global coordinate system into its corresponding frame $\{1\}$
\begin{align}
    \pmb b_{2}^{(1)} & =  \pmb A_{1}^{-1}(\pmb b_{2} - \pmb b_{1}),& \pmb A_{2}^{(1)} & = \pmb A_{1}^{-1} \pmb A_{2},\\
    \hat{\pmb b}_{2}^{(1)} & =  \hat{\pmb A}_{1}^{-1}(\hat{\pmb b}_{2} - \hat{\pmb b}_{1}),& \hat{\pmb A}_{2}^{(1)} & = \hat{\pmb A}_{1}^{-1} \hat{\pmb A}_{2}.
\end{align}
The same applies to the demonstration
\begin{equation}
    \pmb \xi_{t}^{(1)} = \pmb A_{1}^{-1}(\pmb \xi_t - \pmb b_{1}).
\end{equation}
Observed from the local reference frame $\{1\}$, the new situation and the situation where the demonstration is performed only differ in the pose of frame $\{2\}$. By doing so, we only need to consider the influence of frame $\{2\}$ on the reference trajectory transformation. This means that we will not need the relevance weights of frame $\{1\}$, allowing us to later construct an objective function that only depends on that of frame $\{2\}$.
\vspace{-4mm}

\subsubsection{Step 2: Frame--weighted Translation.} We then perform a frame--weighted translation transformation on the reference trajectory along the vector that represents the position difference between frames $\{2\}$. Multiplying the vector by the corresponding frame relevance weight, we can transform each data point $\pmb \xi_{t}^{(1)}$ as
\begin{equation} \label{eq: translation}
    \tilde{\pmb \xi}_{t}^{(1)} =  \pmb \xi_{t}^{(1)} + f_2(d_{t}) \, (\hat{\pmb b}_{2}^{(1)} - \pmb b_{2}^{(1)}),
\end{equation}
where $\tilde{\pmb \xi}_{t}^{(1)}$ is the estimated data point based only on the frame position difference, $d_t$ is the relative distance index of the sample $\pmb \xi_{t}^{(1)}$, and $f_2(d_{t})$ represents the relevance weight of frame $\{2\}$ for this sample.
\vspace{-4mm}

\subsubsection{Step 3: Frame--weighted Rotation.} We further transform $\tilde{\pmb \xi}_{t}^{(1)}$ based on the orientation difference between frames $\{2\}$. We determine the orientation of the new frame $\{2\}$ relative to the frame $\{2\}$ in the demonstrated situation through the matrix operation $\hat{\pmb A}_{2}^{(1)} (\pmb A_{2}^{(1)})^{-1}$. It is then represented as a rotation vector $\theta \, \pmb u$, where $\theta$ is the rotation angle and the unit vector $\pmb u$ indicates the rotation axis. Multiplying the angle $\theta$ with the frame relevance weight $f_2(d_{t})$, we obtain the required rotation vector for transformation as $f_2(d_{t}) \, \theta \, \pmb u$. After converting it back to the rotation matrix $\pmb T_t$, we get the final estimated data point as
\begin{equation}  \label{eq: rotation}
   \hat{\pmb \xi}_{t}^{(1)} =  \pmb T_t (\tilde{\pmb \xi}_{t}^{(1)} - \hat{\pmb b}_{2}^{(1)})+ \hat{\pmb b}_{2}^{(1)},
\end{equation}
where we rotate the previously estimated point $\tilde{\pmb \xi}_{t}^{(1)}$ around the origin of the new frame $\{2\}$, thus aligning trajectory points with its orientation.
\vspace{-4mm}

\subsubsection{Step 4: Projection to Global.} Finally, we obtain the desired trajectory $\{ \hat{\pmb \xi}_{t}\}_{t=0}^{T}$ in the new situation by projecting the local data point $\hat{\pmb \xi}_{t}^{(1)}$ back to the global coordinate system as
\begin{equation}
    \hat{\pmb \xi}_{t} =  \hat{\pmb A}_1 \hat{\pmb \xi}_t^{(1)} + \hat{\pmb b}_{1}.
\end{equation}

\subsection{Frame Relevance Weights Optimization} \label{sec: frame relevance weights optimization}
Since we assume the task trajectories in the same situation are similar, a suitable objective function can be defined between expert demonstrations and synthetic trajectories generated through reference trajectory transformation. The objective function is designed to guide the transformation process, aiming to replicate the expert data. We then formulate the optimization problem as
\begin{equation} \label{eq: final optimization}
\begin{aligned}
    \min_{\pmb{\omega}_2} \quad &  \frac{1}{M(M-1)} \sum_{i=1}^{M} \sum_{\substack{k=1 \\ k \neq i}}^{M} L( \pmb X_i, \ \hat{\pmb X}_{i, k})\\
    \textrm{s.t.} \quad & 0 \leq f_2 \leq 1 ,
\end{aligned}
\end{equation}
where $\hat{\pmb X}_{i, k}$ is the reproduction of the demonstration $\pmb X_i$ by transforming the reference trajectory $\pmb X_k$ and $L(\cdot)$ defines the trajectory dissimilarity as their normalized distance computed using dynamic time warping (DTW) \cite{muller_dynamic_2007}. We find the optimal basis function weights for frame relevance weights by achieving the best match between demonstrated trajectories and their reproductions. Given a training dataset with $M$ demonstrations, we can find up to $M(M-1)$ pairs of demonstrations for trajectory reproduction. Popular optimization algorithms, such as Sequential Least Squares Programming (SLSQP), can be used to solve this problem.

\subsection{Skill Generalization} \label{sec: skill generalization}
Once we have determined the optimal frame relevance weights, we can utilize them for the generalization in a new situation through the reference trajectory transformation described in Sect. \ref{sec: reference trajectory transformation}, which forms our frame--weighted motion generation method. Note that the optimal frame relevance weights are found by considering all demonstrations in the training dataset, so any demonstration can be used as a reference trajectory in theory. In addition, the trajectories generated through our method can serve as valuable additions to augment the original training dataset for traditional TP-LfD methods, as we will later showcase using TP-GMM.

\subsection{Multi--Frame Task Solution} \label{sec: multi-frame task solution}
To extend the application of our proposed method to more complex tasks involving multiple reference frames, we can divide the whole task into a sequence of two-frame subtasks based on our prior knowledge, e.g., change in contact force \cite{kober_learning_2015} and gripper closure \cite{meszaros_learning_2022}. Note that the demonstrations are still collected for the entire task, but we later segment them into individual trajectories modulated by only two reference frames. After employing our method for each movement segment, we can connect the generalized trajectories of each subtask to form a complete one for the whole task\footnote{Refer to \href{https://youtu.be/JpGjk4eKC3o}{the experiment video} for the application to a three-frame task.}.

\section{Experiments} \label{sec: experiments}
\begin{figure}[!t]
    \centering
    \subfloat[Two Franka-Emika Panda robots]{
    \includegraphics[height=0.152\textwidth]{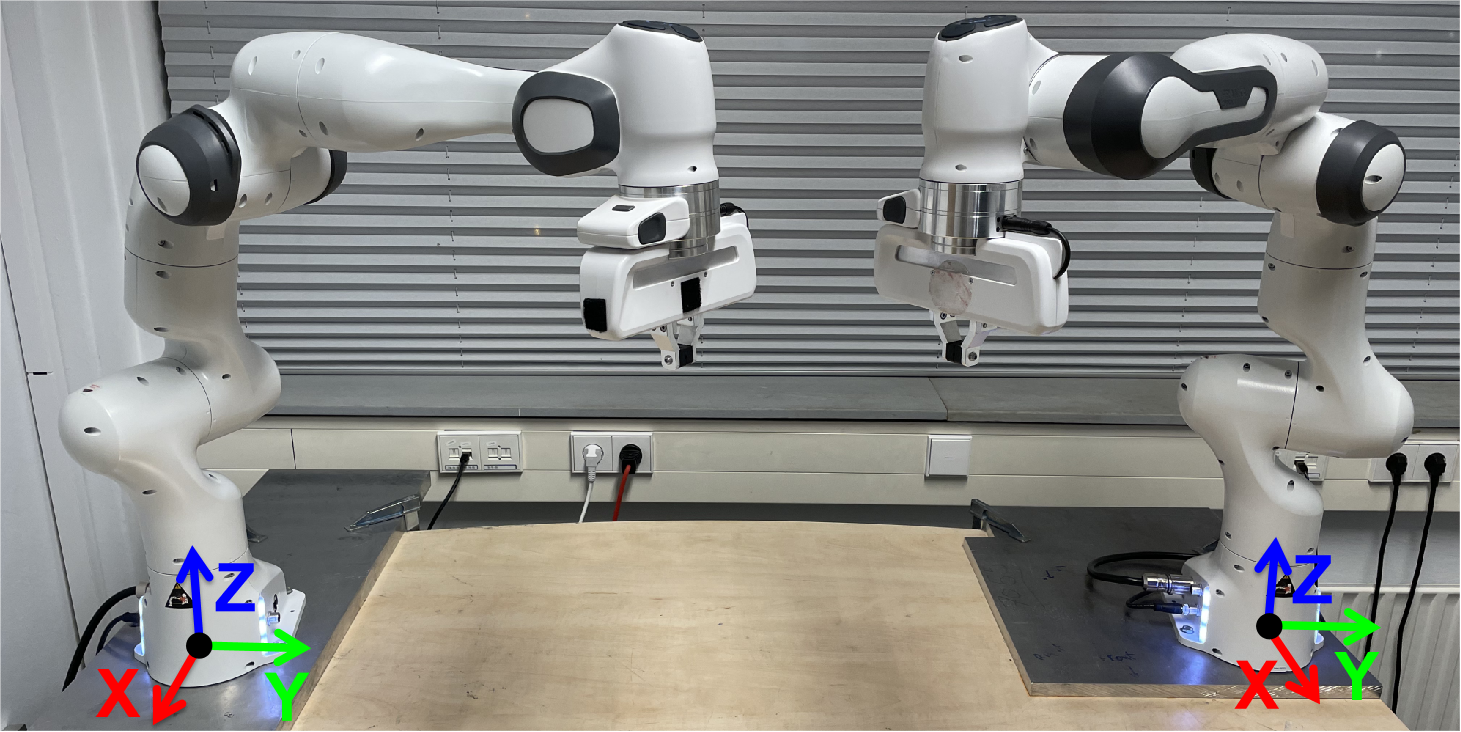}
    \label{fig: bimanual setup}}
    \hfill
    \subfloat[Roller-on-holder task: initial and final state]{
    \includegraphics[height=0.152\textwidth]{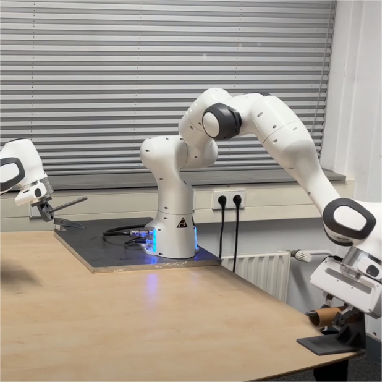}
    \includegraphics[height=0.152\textwidth]{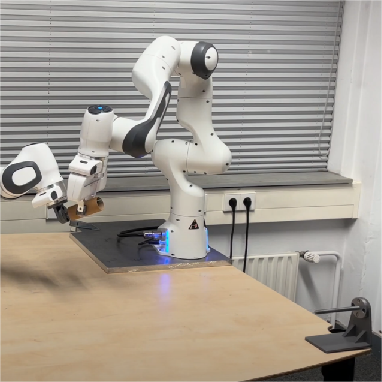}
    \label{fig: roller}}
    \hfill
    \subfloat[Flower-in-vase task: initial and final state]{
    \includegraphics[height=0.152\textwidth]{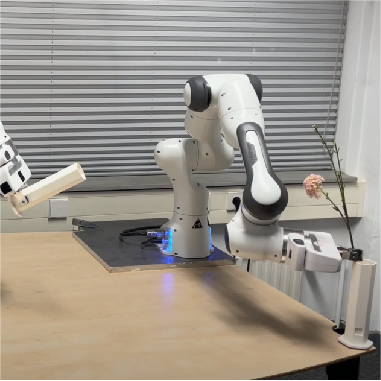}
    \includegraphics[height=0.152\textwidth]{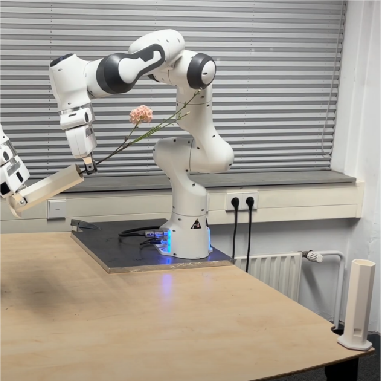}
    \label{fig: flower}}

    \caption{The bimanual setup of robot experiments and the tasks.}
    \label{fig: experiments}
\end{figure}

In our experimental setup, we employ a bimanual configuration consisting of two 7 DoF Franka-Emika Panda robots in the same base orientation, see Fig. \ref{fig: bimanual setup}. The impedance controllers are deployed on both manipulators \cite{franzese_interactive_2023}. We perform two experiments with the real robot setup: \textit{i}) a roller-on-holder task, where the right robot picks the paper roller from the holder on the table and places it onto the other holder mounted on the left robot, see Fig. \ref{fig: roller}, \textit{ii}) a flower-in-vase task, where the right robot grasps the flower from the vase on the table and transfers it to the other vase held by the left robot, see Fig. \ref{fig: flower}. The situation of both tasks can be fully described by two reference frames attached to the starting and ending holder or vase. 

For both experiments, the left robot provides the pose of the ending holder or vase while the right one performs the task. During task executions, we employ Spherical Linear Interpolation (Slerp) to enable the right robot to uniformly adjust its end-effector orientation until the grasped roller or flower is aligned with the ending holder or vase \cite{shoemake_animating_1985}. We then focus on learning the 3D XYZ position of trajectories.

For each task, we use 2 expert demonstrations for training and 4 for validation. We conduct a comprehensive evaluation of our proposed frame--weighted motion generation method on skill generalization by comparing it with TP-GMM and augmented TP-GMM. The latter trains TP-GMM on a training dataset that is augmented with the trajectories generated by our method. We set the number of Gaussian components in TP-GMM as 6 and augment the original training dataset to 9 demonstrations for augmented TP-GMM. To quantitatively compare the three motion generalization methods, we define the model error as the average of the DTW normalized distance between ground truths and generalized trajectories
\begin{equation} \label{eq: error}
    \mathcal{J}_\mathrm{DTW} = \frac{1}{\mu} \sum_{i=1}^\mu L (\pmb X_i, \hat{\pmb X}_i ), 
\end{equation}
where $\pmb X_i$ is the demonstrated trajectory, $\hat{\pmb X}_i$ is the trajectory produced by models, and $\mu$ is the number of trials for generalization (2 for training and 4 for validation).

\section{Experimental Results \& Insights} \label{sec: insights}
\begin{table}[t] 
    \centering
    \caption{Training and validation errors for two real robotic tasks}
    \scalebox{0.74}{
    \begin{tabular}{c|c|c|c|c|c|c}
        \hline
        & \multicolumn{3}{c}{Roller-on-holder task} & \multicolumn{3}{|c}{Flower-in-vase task}\\
        \cline{2-7}
        & \makecell[c]{Frame--weighted \\ motion generation} & \makecell[c]{Augmented \\ TP-GMM} & TP-GMM & \makecell[c]{Frame--weighted \\ motion generation} & \makecell[c]{Augmented \\ TP-GMM} & TP-GMM \\
        \hline
        Training error & \textcolor[rgb]{0, 0.8, 0}{0.0151 (52\%)} & \textcolor[rgb]{0.9, 0, 0}{0.0393 (134\%)} & 0.0293 (100\%) & \textcolor[rgb]{0, 0.8, 0}{0.0214 (50\%)} & \textcolor[rgb]{0.9, 0, 0}{0.0591 (138\%)} & 0.0427 (100\%)\\
        \hline
        Validation error & \textcolor[rgb]{0, 0.8, 0}{0.0805 (12\%)} & \textcolor[rgb]{0, 0.8, 0}{0.1687 (25\%)} & 0.6657 (100\%) & \textcolor[rgb]{0, 0.8, 0}{0.0786 (11\%)} & \textcolor[rgb]{0, 0.8, 0}{0.1639 (23\%)} & 0.7062 (100\%)\\
        \hline
    \end{tabular}}
    \label{tab: exp comparison}
\end{table}

The training and validation errors for both tasks are listed in Table \ref{tab: exp comparison}. While the former indicates the model performance in performing reproduction under observed conditions, the latter represents the ability to generalize to unseen situations. We use the error of TP-GMM as the reference value and express each error value as a percentage relative to this reference. A larger error is marked with red, and a smaller one with green. 

\begin{floatingfigure}[r]{0.3\textwidth}
    \centering
    \includegraphics[width=0.3\textwidth]{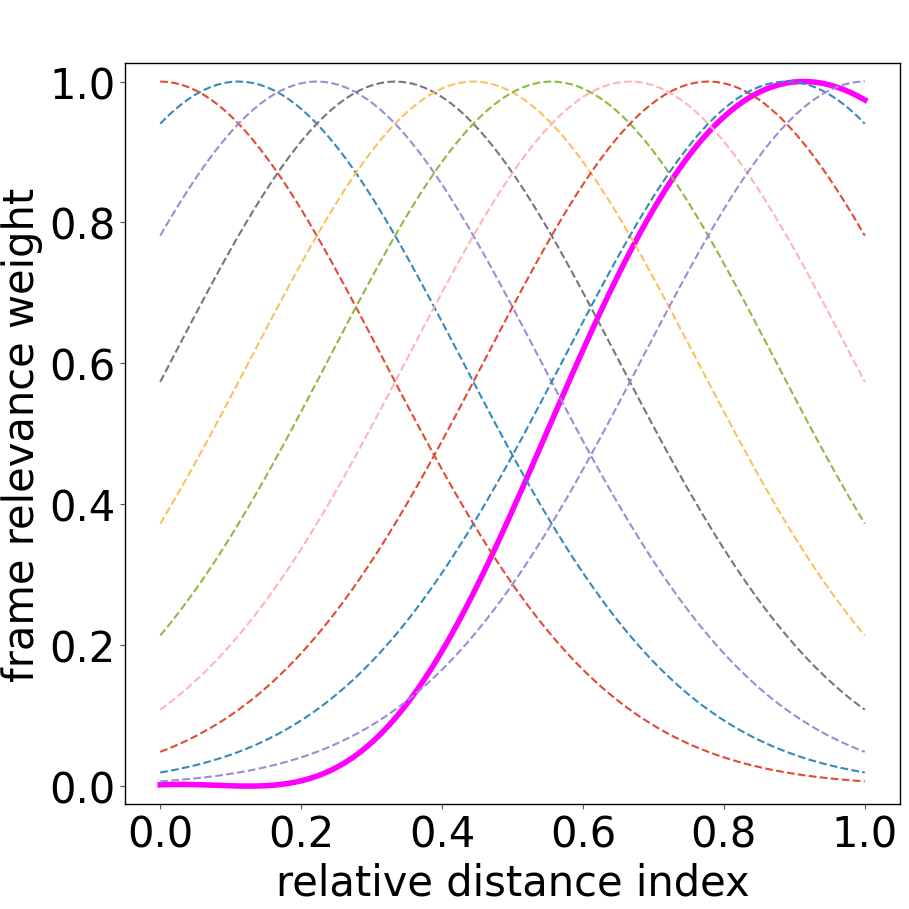}
    \caption{The optimized relevance weights of frame $\{2\}$ (magenta solid line) with RBFs (dashed lines) for the flower-in-vase task.}
    \label{fig: flower weight}
\end{floatingfigure}

Among the three methods, the frame--weighted motion generation method incurs the lowest error for both training and validation. This underscores the pivotal role of frame relevance, which captures how reference frames modulate task trajectories, in the process of skill acquisition. It also highlights the effectiveness of our proposed method in determining it, even with only few demonstrations available.

Comparing the model error of TP-GMM and augmented TP-GMM, we observe that when the original training dataset is augmented with trajectories synthesized using our method, the validation error decreases significantly, but the training error increases slightly. It suggests that TP-GMM, when trained with a limited number of demonstrations, tends to overfit the few and sparse data, thus showing poor performance in terms of generalization to new situations. In addition, it proves that our proposed method is valid in data augmentation.

Taking the flower-in-vase task as an example, we will now present a series of more intuitive results. Figure \ref{fig: flower weight} displays the optimized relevance weights of frame $\{2\}$. To approximate them, we use 10 RBFs with centers uniformly distributed between 0 and 1 and spreads set to 5. Aligned with our expectations, the relevance weights of frame $\{2\}$ gradually increase as data points get closer to the ending vase and finally remain near 1 due to its geometric constraints.

\begin{figure}[!t]
    \begin{minipage}[!t]{0.6\textwidth}
        \centering
        \rotatebox{90}{\scriptsize situation 1}
        \includegraphics[width=0.29\textwidth]{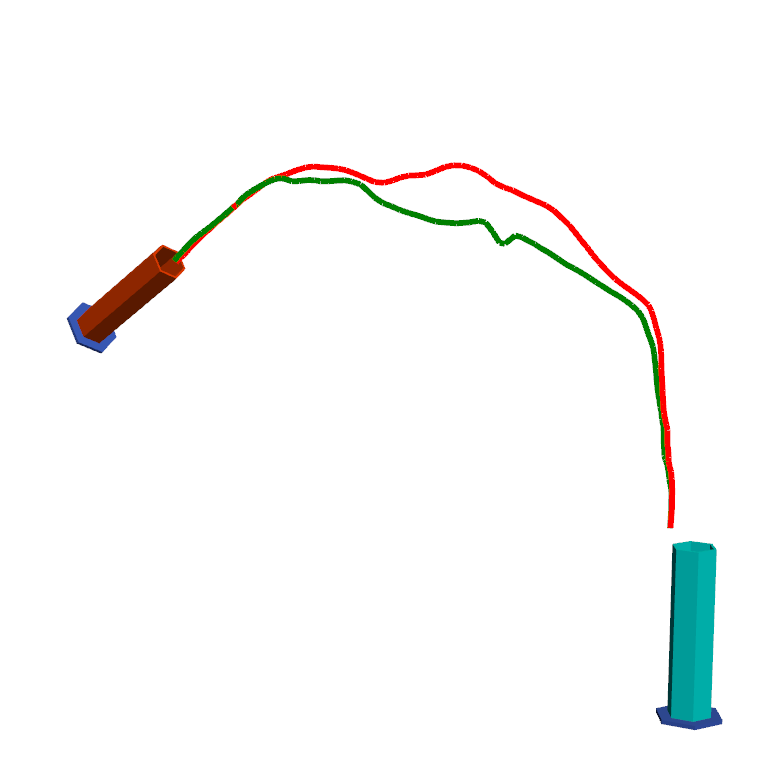}
        \label{fig: flower fwbil validation 1}
        \hfil
        \includegraphics[width=0.29\textwidth]{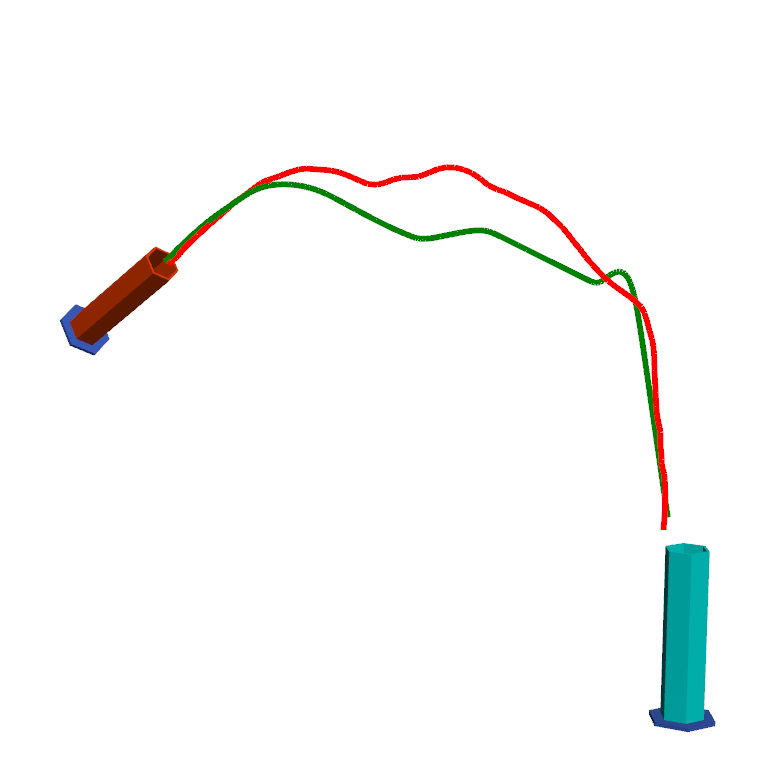}
        \label{fig: flower a-tpil validation 1}
        \hfil
        \includegraphics[width=0.29\textwidth]{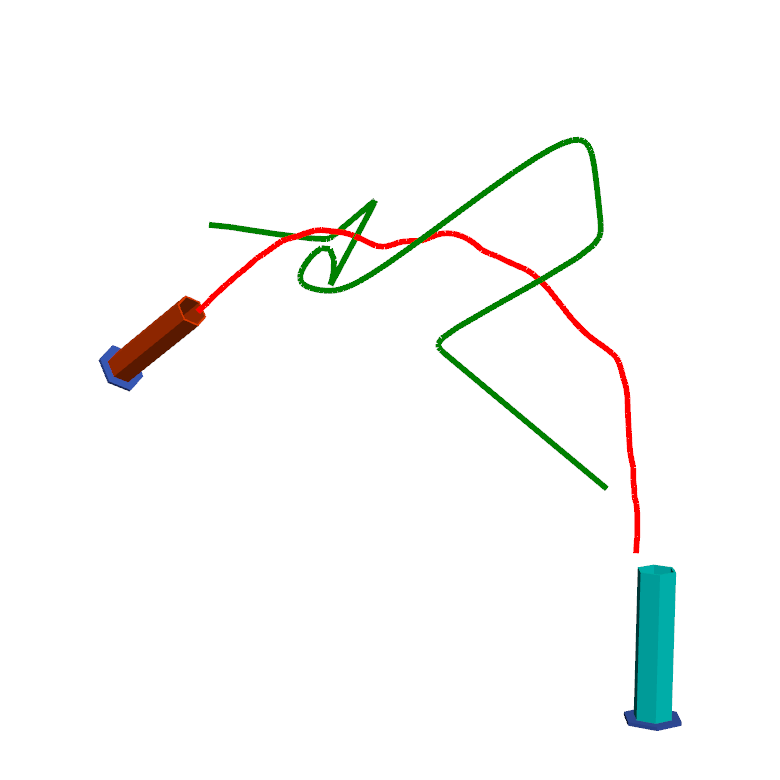}
        \label{fig: flower tpil validation 1}
    
        \vspace{-0.5cm}
        \rotatebox{90}{\scriptsize situation 2}
        \subfloat[Frame--weighted motion generation]{\includegraphics[width=0.29\textwidth]{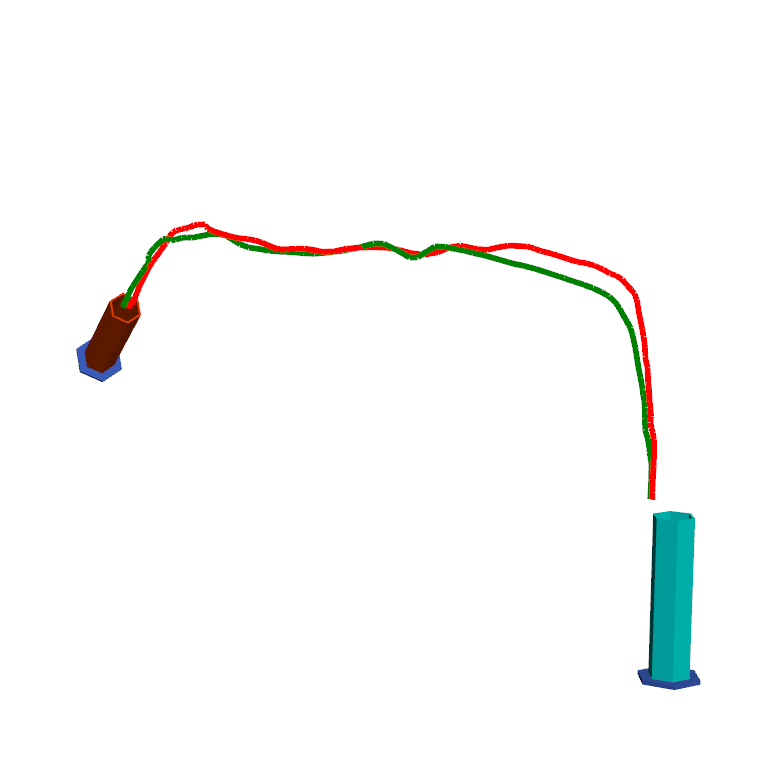}
        \label{fig: flower fwbil validation 2}}
        \hfil
        \subfloat[Augmented TP-GMM]{\includegraphics[width=0.29\textwidth]{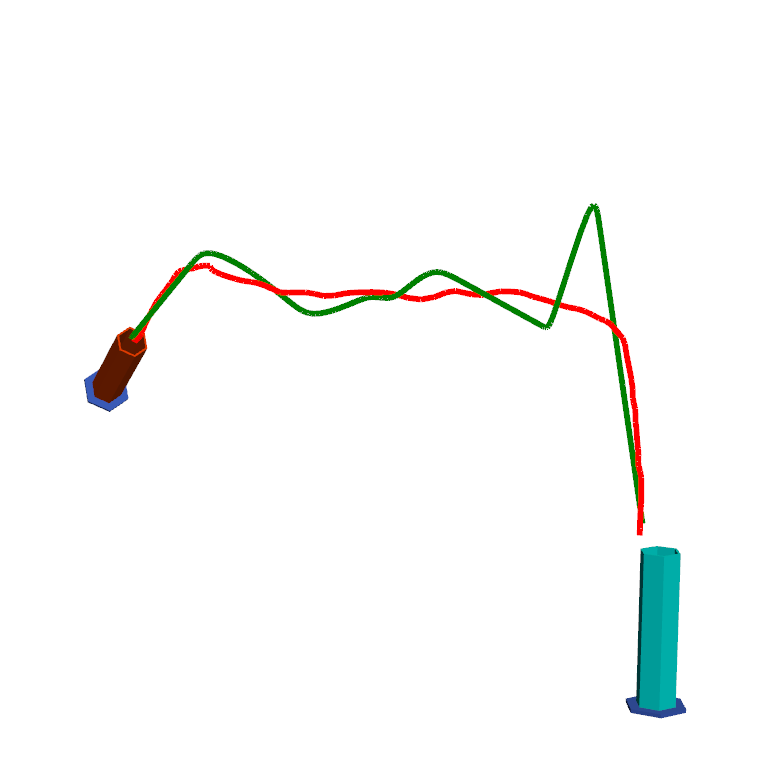}
        \label{fig: flower a-tpil validation 2}}
        \hfil
        \subfloat[TP-GMM]{\includegraphics[width=0.29\textwidth]{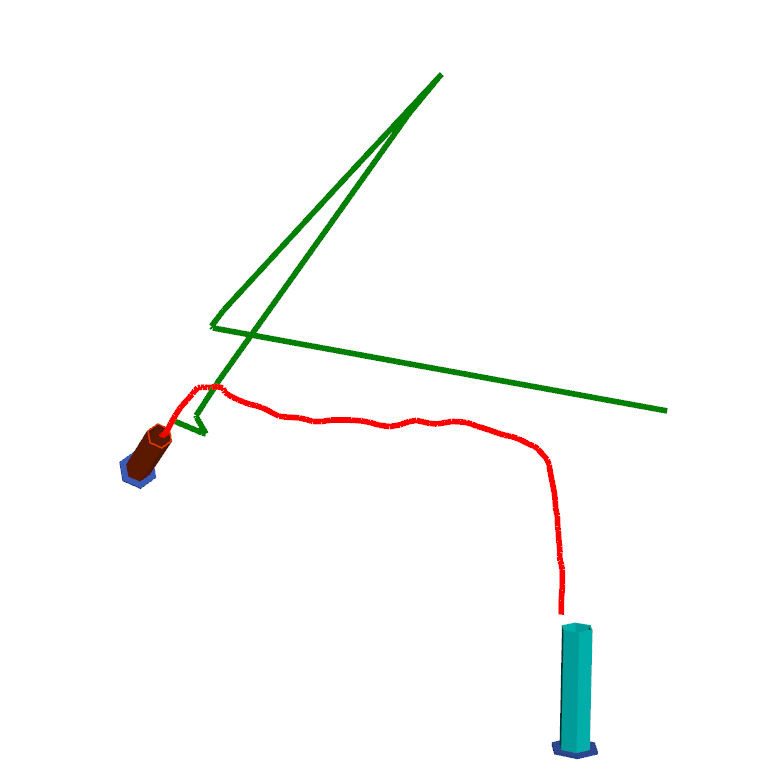}
        \label{fig: flower tpil validation 2}}
        \caption{The generalization performance of three methods on two validation situations in the flower-in-vase task. The starting and ending vases are shown in cyan and red, the generalized trajectories and ground truths in green and red.}
        \label{fig: flower validation}
    \end{minipage}
    \hfil
    \begin{minipage}[!t]{0.35\textwidth}
        \centering
        \includegraphics[width=0.97\textwidth]{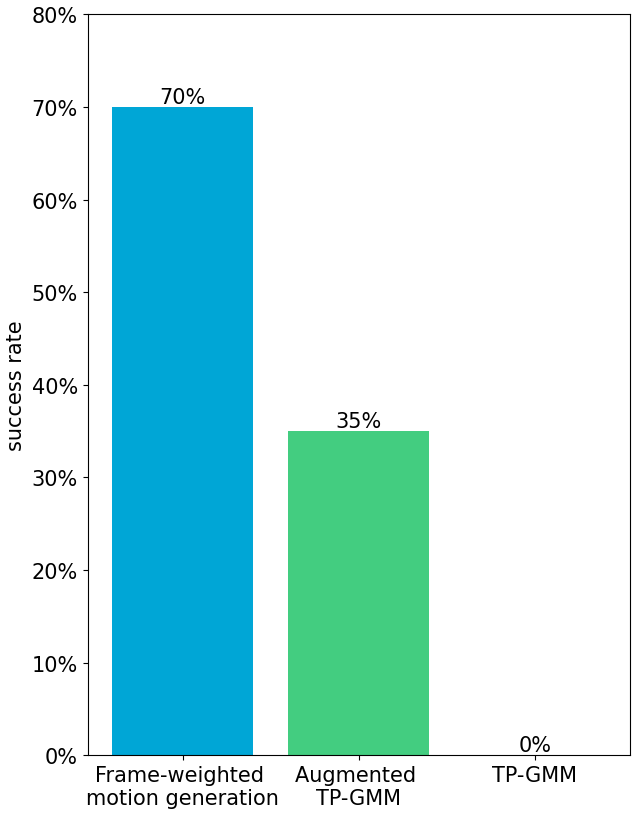}
        \caption{The success rate of three methods on the flower-in-vase task across 20 randomly generated situations.}
        \label{fig: success rate}
    \end{minipage}
\end{figure}

Figure \ref{fig: flower validation} presents how the three motion generalization methods perform in two validation situations. With only two training demonstrations, TP-GMM lacks the statistical basis to model the skill, resulting in meaningless trajectories, as shown in Fig. \ref{fig: flower tpil validation 2}. When the original training dataset is augmented, the generalization performance of TP-GMM improves a lot. Although the two generalized trajectories in Fig. \ref{fig: flower a-tpil validation 2} exhibit some unwanted movements, they are successful in task executions since the geometric constraints of the vase have been satisfied. Figure \ref{fig: flower fwbil validation 2} demonstrates the remarkable similarity between the trajectory produced using our method and the ground truth. While fulfilling the task constraints, these generated trajectories are much smoother.

To further evaluate and compare their generalization capabilities, we test them in a set of 20 new situations, where the poses of both vases are randomly generated within the kinematic limits of the task. If the flower can be grasped from the source vase and placed into the target vase without any collision with either vase, the task is a success. The success rates (percentage of success out of 20 situations) for these three methods are summarized in Fig. \ref{fig: success rate}. The results are consistent with our previous findings. Without enough data for training, TP-GMM fails to generalize in all cases. While augmented TP-GMM achieves a success rate of 35\%, our method has twice the success rate, at 70\%. This underscores its capacity for effective extrapolation, particularly in situations that have a significant difference from what is observed during training.

\begin{figure}[!t]
    \centering
    \subfloat[More training demonstrations]{\includegraphics[width=0.512\textwidth]{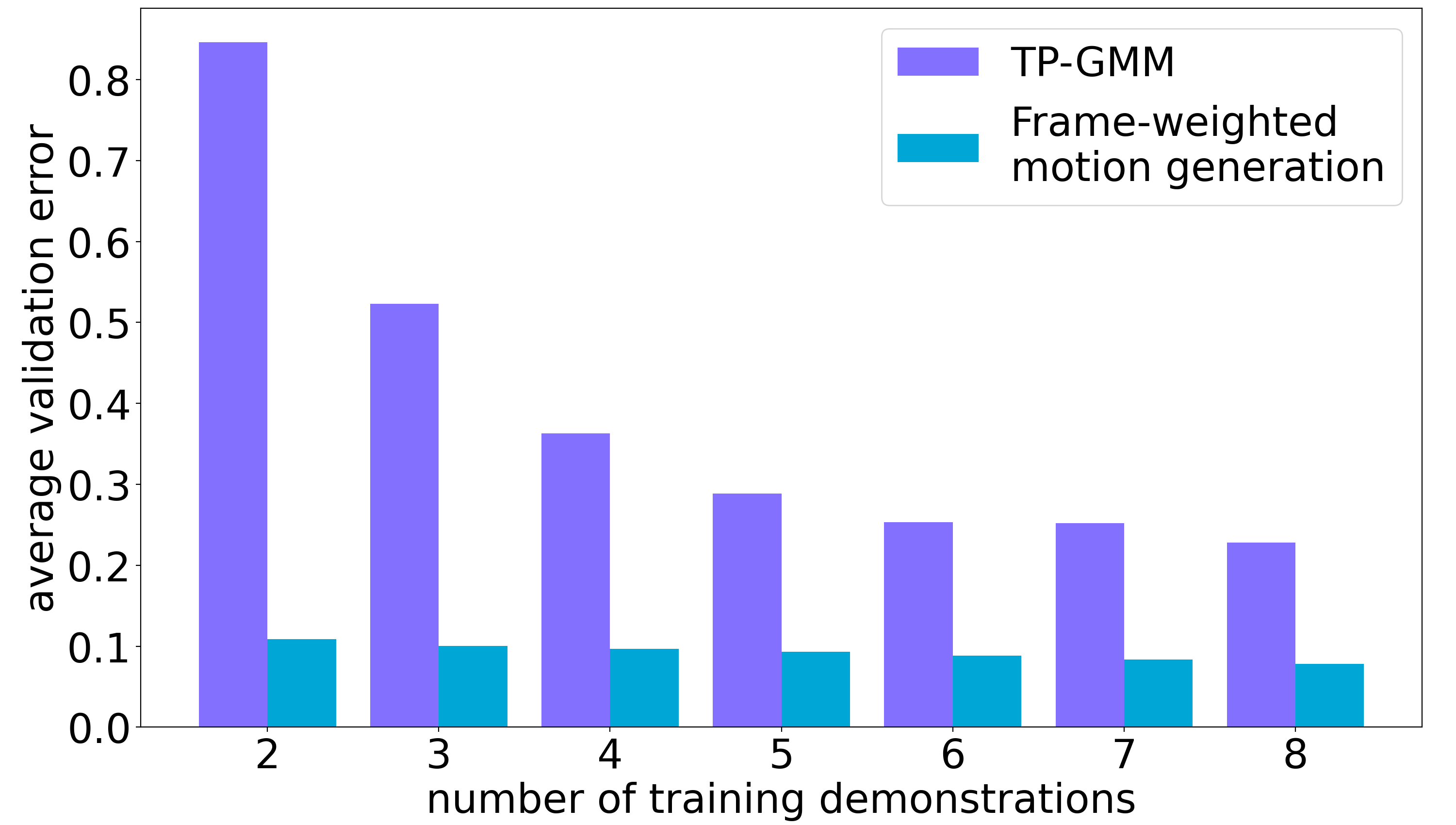}
    \label{fig: tpil and fwbil with more demos}}
    \hfil
    \subfloat[More augmentations]{\includegraphics[width=0.408\textwidth]{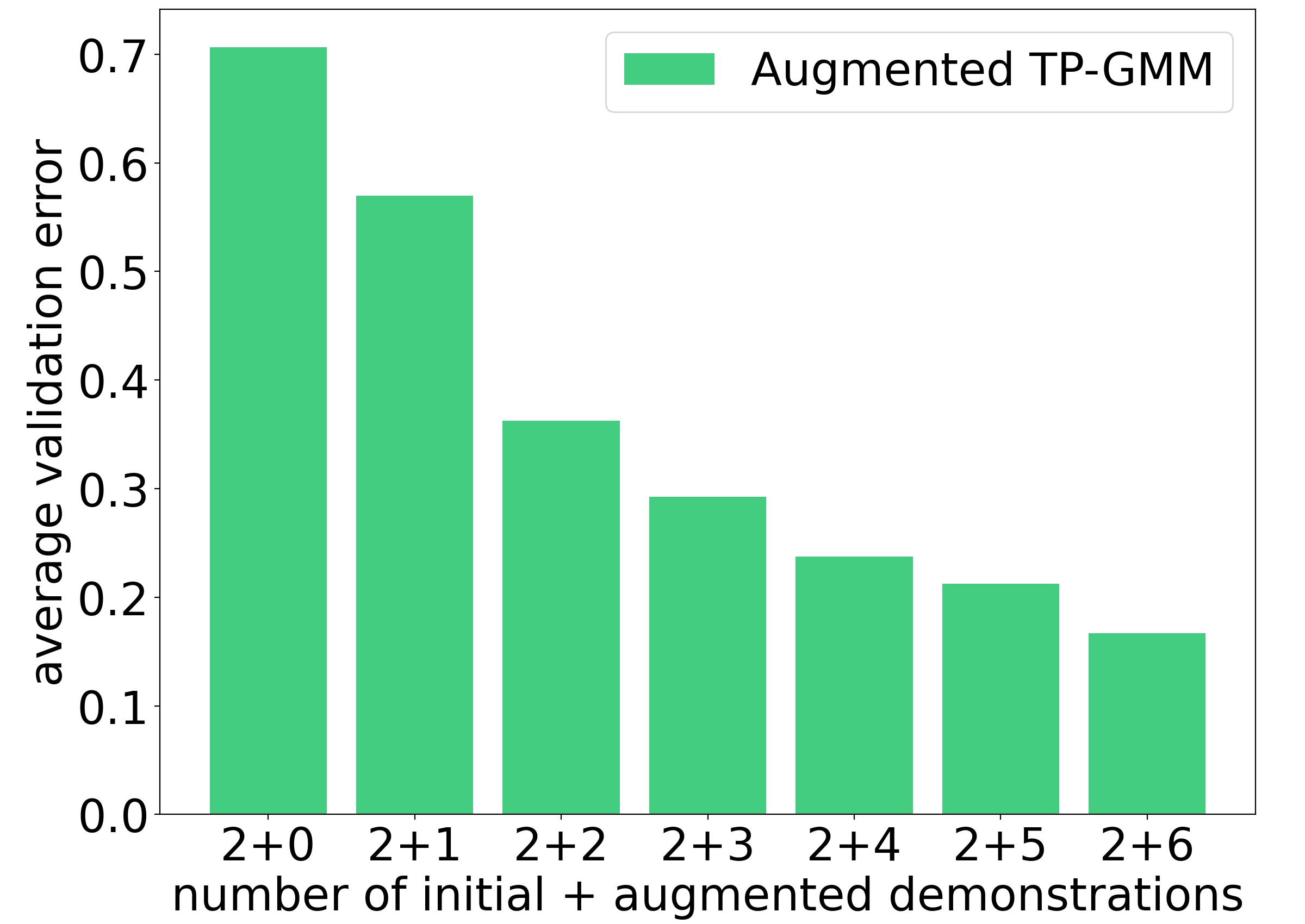}
    \label{fig: augmented tpil with more augmentations}}
    
    \caption{The average validation error and the number of demonstrations in the flower-in-vase task.}
    \label{fig: more demos}
\end{figure}

Figure \ref{fig: tpil and fwbil with more demos} shows how the average validation error changes with an increasing number of training demonstrations. Here, the average means that for each number, we select various combinations of demonstrations for training and calculate the mean error across these combinations. As the number of training demonstrations increases, the error of TP-GMM drops a lot. In contrast, the error of our method decreases slightly, which can be explained by the fact that the frame relevance, an inherent feature of reference frames, does not change with the size of the training dataset. It also means that our method can find the optimal frame relevance weights even with limited data availability. The comparison indicates that our proposed method has less dependence on data, which further validates its ability to learn skills from few demonstrations.

To better demonstrate the validity of our proposed method in data augmentation, Fig. \ref{fig: augmented tpil with more augmentations} shows how the average validation error of augmented TP-GMM changes with an increasing number of augmented demonstrations produced by the frame--weighted motion generation method. For each number, we consider multiple sets of synthetic trajectories for augmentation. The average is calculated by taking the mean of resulting errors. We can observe that the synthetic data generated using our method helps to significantly reduce the model errors, highlighting its value in policy improvement for TP-GMM.

\section{Conclusion} \label{sec: conclusion}
This paper presents a method to learn skills from few demonstrations by building on the calculation of frame relevance weights. Our method reduces the reliance on data while demonstrating excellent generalization performance. We view this capability as an important step towards improving data efficiency in LfD.

Our proposed method is mainly limited by its ability to deal with tasks that require multiple reference frames to describe situations. The segmentation of demonstrations into movement segments introduces extra time and effort and relies on prior knowledge. The next step would be exploring how to extend our method for direct application to multi-frame tasks or achieve automatic segmentation of human demonstrations with limited data. In addition, our method focuses on learning trajectory positions, limiting its applicability in scenarios where the end-effector orientation matters for the success of tasks. For example, specific end-effector rotations may be required to navigate around obstacles effectively. Further research should explore the comprehensive acquisition of both position and orientation information.

\paragraph{Acknowledgements.} This work was supported by Honda Research Institute Europe GmbH as part of the project ``Learning Physical Human-Robot Cooperation Tasks'' and has been partially funded by the ERC Starting Grant TERI ``Teaching Robots Interactively'', project reference \#804907.

\bibliographystyle{styles/bibtex/spmpsci}
\bibliography{reference}

\end{document}